\documentclass[journal,twoside,web]{ieeecolor}
\usepackage{tmi}
\usepackage{cite}
\usepackage{amsmath,amssymb,amsfonts}
\usepackage{algorithmic}
\usepackage{graphicx}
\usepackage{textcomp}
\usepackage{bbding}
\usepackage[misc]{ifsym}
\usepackage{float}
\usepackage{color}
\usepackage[ruled]{algorithm2e}
\usepackage{caption}
\usepackage{subfig}
  \usepackage[switch]{lineno}
\usepackage{url}
\usepackage{booktabs}
\usepackage{multirow}
\usepackage{mathrsfs}
\usepackage{dsfont}
\usepackage{bm}

\newtheorem{definition}{Definition}

\newtheorem{proposition}{\bf Proposition}

\newcommand{\dx}{{\mathrm{d} x}}

\begin{document}
\title{Convex Shape Prior for Deep Neural Convolution Network based Eye Fundus Images Segmentation}

\author{Jun Liu, Xue-Cheng Tai, and Shousheng Luo 
\thanks{This work was supported in part by National Natural Science Foundation of China (No. 11871035), Hong Kong Baptist University through grants RG(R)-RC/17-18/02-MATH, HKBU 12300819, and NSF/RGC grant N-HKBU214-19.}
\thanks{Jun Liu is with School of Mathematical Sciences, Laboratory of Mathematics and
Complex Systems, Beijing Normal University, Beijing 100875, P.R. China. (e-mail: jliu@bnu.edu.cn).}
\thanks{Xue-cheng Tai is with the Department of Mathematics, Hong Kong Baptist University, Kowloon Tong, Hong Kong. (e-mail: xuechengtai@hkbu.edu.hk).}
\thanks{Shousheng Luo is with School of Mathematics and Statistics, Henan University, Kaifeng, P.R. China. (e-mail: sluo@henu.edu.cn).}}
\maketitle

\begin{abstract}
Convex Shapes (CS) are common priors 
for optic disc and cup segmentation in eye fundus images. It is important to design proper techniques to represent convex shapes. So far, it is still a problem to guarantee that the output objects from a Deep Neural Convolution Networks (DCNN) are convex shapes. In this work, we propose a technique which can be easily integrated into the commonly used DCNNs for image segmentation and guarantee that outputs are convex shapes. This method is flexible and it can handle multiple objects and allow some of the objects to be convex.  Our method is based  on the dual representation of the sigmoid activation function in DCNNs. In the dual space, the convex shape prior can be guaranteed by a simple quadratic constraint on a binary representation of the shapes. Moreover, our method can also integrate spatial regularization and some other shape prior using a soft thresholding dynamics (STD) method.  The regularization can make the boundary curves of the segmentation objects to be simultaneously smooth and convex. We design a very stable active set projection algorithm to numerically solve our model. This algorithm can form a new plug-and-play DCNN layer called CS-STD whose outputs must be a nearly binary segmentation of convex objects. In the CS-STD block,
the convexity information can be propagated to guide the DCNN in both forward and backward propagation during training and prediction process.  As an application example, we apply the convexity prior layer
to the retinal fundus images segmentation by taking the popular DeepLabV3+ as a backbone network.
Experimental results on several public datasets show that our method is efficient and  outperforms the classical DCNN segmentation methods.
\end{abstract}


\section{Introduction}
\IEEEPARstart{C}onvex shapes are common in our daily life. For instance, buildings, cars, balls, books are all convex objects we need to handle often.
Many organs and tissues are convex in medical images. Taking an example, the optic disc and cup in eye fundus images are always convex. The optic disc formation is a sensitive factor in glaucoma. The vertical cup to disc ratio
plays a key role in early diagnosis of glaucoma. Thus, to accurately segment the optic disc and cup for retinal images can help diagnosis and treatment of glaucoma. There is some general knowledge for optic disc and cup regions. First, the optic disc always fully contains the cup. Secondly,
the boundaries of the disc and cup are smooth. Thirdly,
these two regions are convex. For a good segmentation algorithm, these priors should be considered. In this paper, we shall develop a deep learning based segmentation method that can easily handle all these spatial priors. Up to now, it is still a problem for commonly used DCNNs to incorporate these kind of spatial priors. 

Still taking the the optic disc and cup in retinal images as an example, to get the segmentation of disc and cup regions, two kinds of methods can be applied. One of the classical image segmentation methods is the handcraft designed algorithm, 
in which a label function is usually used to represent different classes of segmentation. In the discrete case, such a label function based multi-phase segmentation model can be exactly solved by the graph cut method \cite{Ishikawa2003} since the energy is submodular \cite{Kolmogorov2004} when the regularization term is anisotropic Total Variation (TV) \cite{Rudin1992}. With a level set formulation \cite{chan2001active}, this method can be extended as the piecewise constant level set method (PCLSM) \cite{Lie2006}. However, the label function based models are not convex and they may have local minimums which are related to undesirable segmentation results. To address this issue, the functional lifting method \cite{Pock2008} can be applied to make the segmentation energy convex with a sublevel set representation. To force the segmentation to be a level function of convex object, several attempts have be done in the references. For the binary classification, Royer \emph{et.al.} \cite{Royer2016} proposed a multi-cut problem for image segmentation with convex shapes according to the
definition of the convex set, i.e.  the line segment between any two points in it should not pass through the object boundary. Another graph cut method is to prevent 1-0-1 configurations in the inner of the convex object area \cite{Gorelick2017}. In \cite{gorelick2017multi},  this method is 
extended to handle multiple convex objects segmentation
% problem was addressed by this method. 
However, these discrete graph cut methods often suffer from measure and mesh errors such that they cannot solve the isotropic TV. The reason for it is that the related energy is not submodular. This drawback would lead to some zigzag edges in the segmentation boundaries \cite{Liu2014}.
 For the continuous segmentation method, convex regions can be guaranteed convexity if the curvature of continuous boundary curve is non-negative  \cite{Yuan2013,yangA2017}. Let $\phi$ be a signed distance function of an object. Then in the level set
method, a simple linear constraint $\triangle \phi\geqslant 0$  for the signed distance function $\phi$ can force the segmented
region to be convex \cite{Yan2018}. This is because the curvature $\kappa=\text{div}(\frac{\nabla\phi}{|\nabla\phi|})$ in the implicit representation of the curves would be reduced to $\kappa=\triangle \phi$ when $|\nabla\phi|=1$ in terms of the proposition of signed distance function. It was extended to multi-object segmentation using a single level set function in \cite{Luo2019}. However, to strictly keep the constraints $|\nabla\phi|=1$ needs to solve a nonlinear PDE, which is
time-consuming. Another drawback of this method is that
it cannot represent multiple connected domains in which all connected components are convex. For convex shape prior with binary segmentation, a quadratic convex shape constraint is proposed in \cite{Luo2020} with discrete curvature $\kappa\geqslant 0$ derived from thresholding dynamics (TD). This relaxed condition can ensure every objects in a multiple connected domains are all convex. However,
compared with the linear constraint in level set method, this nonlinear condition is more difficult to numerically implement. If one adopts the Lagrangian multiplier type method (e.g. \cite{Luo2020}), it is difficult to choose the step size of iteration and the algorithm may converge slowly and unstably sometimes. In this paper, we will develop a very stable and efficient projection algorithm to numerically keep this convex shape condition.

One main drawback of the handcraft designed model based segmentation is that the features used for classification are always manually selected. It fails to extract some complicated low level and group features when given many supervised sample pairs. On the other hand, the DCNNs based segmentation can extract nonlinear deep features and they can produce promising results on big datasets.
The DCNNs based learning method has produced very successful results for image segmentation. Since the Fully Convolutional Network (FCN) \cite{Long2014Fully} was proposed, the encoder-decoder network architecture has became a standard
paradigm. The representative encoder-decoder networks are U-net \cite{Ronneberger2015} and DeepLab series \cite{v1,v2,v3+}. Recently, many variants such as V-net \cite{milletari2016v}, M-net\cite{Fu2018}, SegNet \cite{SegNet} have been proposed for image segmentation.
More deep learning methods for image segmentation can be found in a recently survey \cite{Minaee2020}. In a general segmentation DCNN, the pooling and dilatation convolutions in the encoder structure can enlarge the reception field and
discover multi-resolution features. However, the spatial position information would be damaged by repeated downsampling and upsampling. Besides, though the DCNNs have strong abilities to extract the deep features for natural images, some basic segmentation requirements, such as spatial smoothness and convex shapes,
cannot be guaranteed. This is because the DCNNs are just continuous mappings, and they fail to describe the specific segmentation spaces such as shape convexity. Moreover,
the classification functions, namely, nonlinear activation functions in DCNNs are often given and they lack spatial dependence. Suitable spatial prior information can help DCNNs to restore some lost spatial information, and can improve the performance of the DCNNs segmentation method. To impose the DCNNs to have specific properties, e.g. they should belong to bounded variation (BV) space in which the functions are piece-wise constants, 
three different appraoches can be used. The first one is the post-processing appraoch. 
For example, one can use a DCNN to extract the features and then put them into the fidelity term of a variational model to segment the objects. 
A typical example for such a method is the  Conditional Random Field (CRF)\cite{crf} post-processing.
In the early deeplab method \cite{v1}, CRF is used to improve the smoothness of the segmentation. The flaw of post-processing method is that the model-based algorithms do not join in the back propagation in the training step and the spatial regularization prior information cannot be transmitted into the DCNNs, 
and thus it is hard to correct the misclassification which comes from DCNNs. 
The second approach is to add regularization term into the loss function. In \cite{Liu2018}, TV was introduced into the loss function for image denoising network. A morphology-aware segmentation loss is added into the segmentation DCNN in \cite{Wang2019}. Though the error information can be back propagated according to
the loss function, and the modified loss function can partly improve the results, it is sensitive to
perturbations of the inputs since the prediction process does not contain loss function. 
To combine both advantages of post-processing and loss function methods, in the third method, 
the spatial regularization term such as TV could be designed into the network structure. 
In \cite{Jia2019}, a TV regularized softmax activation function was proposed and it enables to put a model based variational segmentation algorithm as one block of the commonly used network architecture. Very recently, to improve the computational efficiency and stability of TV block in DCNNs, a Soft Thresholding Dynamics (STD) softmax activation function was introduced in \cite{Liu2020}. In this paper, we will show that  the sigmoid activation function also can be regularized and projected on a convex shape space in the similar way. It can guarantee the convexity of the DCNNs outputs.
As far as we known, there is no work on DCNNs architecture to guarantee convexity of  output  shapes for image segmentation.

The objective of this paper is to integrate the convex shape prior in variational segmentation method into the deep learning. We will design a DCNN block to be integrated into commonly used DCNNs and it guarantees that the outputs must have some mathematical properties such as only containing convex objects. 
%Therefore, it can improve the segmentation results when the datasets have convex shape prior.

The main contributions of the study are:
\begin{itemize}
\item We propose a general method to integrate convex shape prior into DCNNs. This is done using a dual formulation of the sigmoid activation function. 
\item A stable and efficient algorithm for keeping a quadratic
convex shape condition is proposed.
%and unrolling the proposed algorithm as a  Convex Shape and Soft Thresholding Dynamics (CS-STD) based sigmoid activation function block for DCNNs.
\item A CS-STD based DeepLabV3+ is proposed to apply the optic disc and cup segmentation for retinal fundus images.
In this application, many techniques in variational based segmentation such as sublevel set representation, spatial regularization, and convex shape are all integrated into DCNNs.
Experimental results show that it can greatly improve the quality of the segmentation results.
\end{itemize}

The rest of this study are organized as following:
The classical model-based segmentation method, convex
shape condition and the DCNNs based deep learning image segmentation methods will be introduced in section \ref{relatedwrok}. Afterwards, we shall propose our method including theory, algorithm, and applications in section \ref{proposed}. Experimental results to evaluate the proposed algorithms are give in section \ref{experiment}.
The final section contains some conclusions and discussions.

\section{Some Related Work}\label{relatedwrok}
\subsection{Some Variational Segmentation Methods}
\subsubsection{Multiple Label Segmentation}
Multiple label model is a classical image segmentation method, and its
minimization problem can be written as
\begin{equation}\label{multi_label}
\underset{l}{\min}\left\{\int_{\Omega} o(v(x),l(x)) \dx+\lambda \int_{\Omega} |\nabla l(x)| \dx\right\},
\end{equation}
where $l:\Omega\rightarrow\{1,\cdots,L\}$ is a label function and $l(x)$ indicates that the pixel located at $x$
belongs to $l(x)$-th class. $o(v(x),l(x))$ is a feature of $l(x)$-th class for a given image $v(x)$. The second total variation (TV) term is to penalize the approximated length of the class boundaries. This model has many variants. For example, in the discrete version, let the regularization be a
discrete anisotropic TV. Then the related segmentation algorithm called Ishikawa graph cut method \cite{Ishikawa2003}. With the level set representation, it is closely associated with the PCLSM \cite{Lie2006}.

One of the main drawbacks of this model is that it is not convex with respect to label function $l$ due to the existence of  complicated  feature $o$. To address this problem, the lifting technique \cite{Pock2008} can be applied to the convexification of multiple label model \eqref{multi_label}. By introducing the $\gamma$-sublevel set functions
\begin{equation}\label{up-lev-set-fun}
u(x,\gamma)=\left\{
\begin{array}{rl}
1,&l(x)\leqslant \gamma,\\
0,&l(x)> \gamma,
\end{array}
\right.
\end{equation}
the energy \eqref{multi_label} becomes a convex one with
respect to the sublevel set $ u$ as follows:
\begin{equation}\label{label_convex}
\underset{ u,\partial_{\gamma}  u\geqslant 0}{\min}\left\{
\begin{array}{r}
\displaystyle\biggl.\int_{\mathbb{R}}\int_{\Omega} o(v(x),\gamma)\partial_{\gamma}u(x,\gamma)\dx\mathrm{d}\gamma\\
+\displaystyle\biggl.\lambda \int_{\mathbb{R}} \int_{\Omega}|\nabla_{x}u(x,\gamma)| \dx \mathrm{d} \gamma
\end{array}
\right\},
\end{equation}
where the label function $l(x)$ and the sublevel set function $ u$ are related by
$$l(x)=l_{max}-\int_{l_{min}}^{l_{max}}u(x,\gamma)\mathrm{d}\gamma.$$
Here $l_{min}$ and $l_{max}$ are the minimum and maximum of $l(x)$, respectively. When $l(x)$ just takes integer values from $1$ to $L$, denote $u(x,\gamma), o(v(x),\gamma)$ as $u_{\gamma},o_{\gamma}$, using the fact $u_L=1$ and the dual representation, and then \eqref{label_convex} is equivalent to
 \begin{equation}\label{label_convex_discrete}
\underset{ \bm u\in\mathbb{U}}{\min}\left\{\sum_{\gamma=1}^{L-1}\int_{\Omega} \left(o_{\gamma}-o_{\gamma+1}\right)u_{\gamma}\dx+\lambda \sum_{\gamma=1}^{L-1} \int_{\Omega}|\nabla u_{\gamma}|\dx \right\},
\end{equation}
where $\mathbb{U}$ is a relaxed sublevel set function set
\begin{equation}\label{sublevelsetu}
 \mathbb{U}=\{\bm u=(u_1,\cdots,u_{L-1}):~~0\leqslant u_{1}(x)\leqslant\cdots\leqslant u_{L-1}(x)\leqslant 1\}.
\end{equation}

The  model \eqref{label_convex_discrete} can segment image into $L$ phases by using $L-1$ binary functions \cite{Liu2014}.
Moreover, the functions in $\mathbb{U}$ are nested, which is beneficial to the retinal images segmentation since the cup and disc regions in eye images are always nested.
When one gets the sublevel set functions $\bm u$, then the label function $l$ can be recovered by formulation
\begin{equation}\label{eq:sublevel2label}
l(x)=L-\sum_{\gamma=1}^{L-1}u_{\gamma}(x).
\end{equation}

%For binary image segmentation problem, it can be solved by
%the well-known Potts model, which minimizes the energy
%\begin{equation}\label{Potts}
%\sum_{l=1}^L\int_{\Omega_l} o_l(x) \dx+\lambda \sum_{l=1}^L |\partial \Omega_l|,
%\end{equation}
%where $o_l(x)$ is a feature of $l$-th class, and $|\partial \Omega_l|$
%is the length of boundary for $l$-th class. $\lambda>0$ is a regularization parameter which controls the balance of the similarity (first) term and regularization (second) term. The binary segmentation condition is $\bigcup_{l=1}^L\Omega_l=\Omega,$ and $ \Omega_{l}\bigcap\Omega_{l^{'}}=\emptyset$ when $l\neq l^{'}$.
%
%With an indicative function of $\Omega_l$, i.e $u_l(x)=1$ if $x\in\Omega_{l}$, else $u_l(x)=0$, then Potts model energy can be a simple formulation
%\begin{equation}
%\mathcal{E}(\bm u)=\underbrace{\langle\bm o, \bm u\rangle}_{:=\mathcal{F}(\bm u)}+\underbrace{\lambda \text{TV}(\bm u)}_{:=\mathcal{R}(\bm u)}.
%\end{equation}

%\subsubsection{Convex Shape Image Segmentation}

\subsubsection{Conditions for Convex Shapes with Binary representation}
To obtain a condition for convex shapes with binary representation, a discrete version of curvature $\kappa\geqslant 0$ for convex objects has been given in \cite{Luo2020}
\begin{proposition}[Convex shape condition \cite{Luo2020}]\label{pro1}
Let $u$ be an indicative function of object region $\mathbb{D}\subseteq \Omega \subseteq \mathbb{R}^n$, i.e. $u(x)=1$ if $x\in\mathbb{D}$, else $u(x)=0$ when $x\in\Omega\cap\mathbb{D}^{c}$. For a given $x\in \Omega$ and $r \ge 0$, let $g_r(x)$ be a kernel function whose support set is a sphere $\mathbb{B}_r=\{x: \|x\|_2\leqslant r\}\subset \Omega$, namely,
\begin{equation*}
g_r(x)=\left\{
\begin{array}{rl}
\frac{1}{|\mathbb{B}_r|},&x\in\mathbb{B}_r \subset \Omega,\\
0,&\text{else}.\\
\end{array}
\right.
\end{equation*}
If $u\in\mathbb{C}$ with $\mathbb{C}$ being defined as 
\begin{equation}\label{eq:C}
\mathbb{C} =\left\{
\begin{array}{r}
u:~~(1-u(x)) (g_r*(1-2u))(x)\geqslant 0, \\
\forall \mathbb{B}_r \subset \Omega, \forall x\in\Omega
\end{array}
\right\},
\end{equation}

then the connected components of $\mathbb{D}$ are all convex.
Here the symbol $*$ stands for the convolution operation and
$\left(g_r*(1-2u)\right)(x)=
\displaystyle\biggl.\int_{\Omega} g_r(x-y)(1-2u(y))\mathrm{d} y$.
\end{proposition}
We want to emphasize that the quadratic constraint in the definition of $\mathbb{C}$ in (\ref{eq:C}) needs to be satisfied for all $r \ge 0$ such that $\mathbb{B}_r \subset \Omega$, not only for one given $r\ge 0$. In our numerical implementations in the discrete setting, we just choose a few values for $r$ and ask the quadratic constraint to be satisfied for these pre-specified values of $r$. 

To numerically keep this quadratic convex shape condition, the gradient descent based Lagrangian multiplier is adopted in \cite{Luo2020}. However, the choice of time step size for gradient descent is difficult and this may lead to the slow convergence of the algorithm sometimes.
In this paper, we propose to use  a very stable and fast algorithm to
numerically keep this convex shape condition.

\subsection{Some Deep Learning Segmentation Methods}
Denote  $\bm v^{0}=\bm v$
as an input of a pixel-wise segmentation DCNN. Then the image segmentation DCNN can be written as a parameterized nonlinear operator $\mathcal{N}_{\bm\Theta}$ defined by $\bm v^T=\mathcal{N}_{\bm\Theta}(\bm v^0)$. The output $\bm v^T$ of the DCNN  is given by the following $T$ layers recursive connections
\begin{equation}\label{eq:nn1}
\left\{
\begin{array}{rl}
\bm o^t=&\mathcal{T}_{\bm \Theta^{t-1}}(\bm v^{t-1},\bm v^{t-2},\cdots,\bm v^0), t=1,\cdots, T,\\
\bm v^{t}=&\mathcal{A}^{t}(\bm o^t), t=1,\cdots, T,\\
\end{array}
\right.
\end{equation}
Here $\mathcal{A}^{t}$ is an activation functional such as the popular ReLU. It also can be downsampling, upsampling operators and their compositions \emph{etc.}. In the last layer,
$\mathcal{A}^{T}$ is a soft classification activation function
such as sigmoid or softmax.
$\mathcal{T}_{\bm \Theta^{t-1}}(\bm v^{t-1},\bm v^{t-2},\cdots,\bm v^0)$ is a given operator which shows the
connections between the $t$-th layer $\bm v^{t}$ and its previous layers $\bm v^{t-1},\bm v^{t-2},\cdots,\bm v^0$.
For the  simplest convolution network, $\bm v^{t}$ is usually only associated to
$\bm v^{t-1}$ and
$\mathcal{T}_{\bm \Theta^{t-1}}(\bm v^{t-1})= \bm w^{t-1} * \bm v^{t-1} + \bm b^{t-1}$ is an affine transformation, in which $\bm w^{t-1}, \bm b^{t-1}$ are convolution kernel and translation, respectively.
$\bm\Theta=\{\bm\Theta^t=(\bm w^{t}, \bm b^t)| t=0,1,\cdots,T-1\}$ is an unknown parameter set.
The output of this network $\bm v^T:\Omega\rightarrow [0,1]^{L}$ should be a soft classification function. For two phases segmentation, it could be sigmoid function. For multi-phase more than 2, the softmax function can be used.
The component function $v^T_{\gamma}(x)$ implies the probability of a pixel located at $x$  belonging to $\gamma$-th class.

By carefully choosing the operator $\mathcal{A}^{t}$ as ReLU, downsampling or upsampling operator, let $\mathcal{T}_{\bm \Theta^{t-1}}$ jump to connect different layers, then the formulation \eqref{eq:nn1} can represent the well-known backbone U-net network \cite{Ronneberger2015}. Similarly, it can be the mathematical formulation of DeepLabV3+ \cite{v3+}.

One may find that the operators $\mathcal{A}^t,\mathcal{T}_{\bm \Theta^{t-1}}$ are continuous or even Lipschitz continuous. Thus $\mathcal{N}_{\bm\Theta}$ is continuous. However, the function space for the output $\bm v^T$
of the DCNN is ambiguous and many existing spatial priors
such as the piece-wise constant proposition and convex object region cannot be guaranteed. To enforce the regularization of $\mathcal{N}_{\bm\Theta}$, the softmax function in the last layer can be replaced by a regularized softmx function through the following variational problem \cite{Jia2019}:
\begin{equation}\label{eq:rnn}
%\left\{
%\begin{array}{rl}
%\bm o^t=&\mathcal{A}^{t}\circ\mathcal{T}_{\bm \Theta^{t-1}}(\bm v^{t-1},\bm v^{t-2},\cdots,\bm v^0),t=1,\cdots, T.\\
%\bm v^{T}=&\arg\min\limits_{\bm u}\left\{\langle-\bm o^T,\bm u\rangle+\varepsilon\langle\bm u,\ln \bm u\rangle\right.
% \left.+\lambda\mathcal{R}(\bm u)\right\}.
%\end{array}
%\right.
\left\{
\begin{array}{l}
\left\{
\begin{array}{rl}
\bm o^t=&\mathcal{T}_{\bm \Theta^{t-1}}(\bm v^{t-1},\bm v^{t-2},\cdots,\bm v^0),t=1,\cdots, T,\\
\bm v^{t}=&\mathcal{A}^{t}(\bm o^t),t=1,\cdots, T-1,\\
\end{array}
\right.
\\
\\
\bm v^{T}=\arg\min\limits_{\bm u}\left\{\langle-\bm o^T,\bm u\rangle+\varepsilon\langle\bm u,\ln \bm u\rangle\right.
\left.+\lambda\mathcal{R}(\bm u)\right\}.\end{array}
\right.
\end{equation}
Here $\varepsilon\geqslant 0$ is an  entropic regularization parameter and $\mathcal{R}$ is a regularization term such as TV.
We can go one step further than the approaches given in \cite{Jia2019,Liu2020}, i.e. we can incorporate convex shape prior into DCNNs by modifying the activation functions.

\section{Our Proposed Method}\label{proposed}
We try to integrate the convex shape prior into DCNNs
through  the dual space of the sigmoid activation function for two phases segmentation. With the sublevel set function representation, multi-phases segmentation can be transformed to several two-phase segmentations. To fit the convex shape prior with binary representation, the smooth sigmoid function would be nearly binary, which can be achieved by setting a small entropic regularization parameter in the dual space.
In addition, we can also incorporate other spatial priors as as convexity of shapes and inclusion of one segmentation region into another one into the DCNNs. 
%to keep the nested connection between cup and
%disc region in eye images, the sublevel set representation in DCNN is adopted.
%\subsection{Upper Level Sets}
%
\subsection{The Dual Representation of Sigmoid Activation Function}
The sigmoid function is usually chosen as the Logistic function $\mathcal{S}(o) (x)=\frac{1}{1+e^{-o(x)}}$. This function can map the feature $o(x)$ from $(-\infty,+\infty)$ to $[0,1]$ to form a probability space. We can easily see that it is spatially independent and the value at $x$ is independent of its neighborhoods. This proposition is not suitable for image segmentation which requires that the segmentation labels are piece-wise constants. In the next, we will show the sigmoid function is a dual function of the smoothed ReLU. Thus the spatial dependence and convex shape prior can be easily added into the dual representation.

Let us recall ReLU$(o)=\max\{o,0\}$. Though $\max$ function is convex, it is not differentiable. We can smooth it with
a $\log$-sum-$\exp$ function.
\begin{definition}[$\log$-sum-$\exp$ function]
\begin{equation*}\label{f_p}
\mathcal{M}_{\varepsilon} (o)=\varepsilon ln( e^{\frac{o}{\varepsilon}}+1).
\end{equation*}
\end{definition}
By some simple calculations, we can get that $\underset{\varepsilon\rightarrow 0}{\lim}\mathcal{M}_{\varepsilon}( o)=\max\{o,0\}$ and $\mathcal{M}_{\varepsilon}( o)$ is convex and smooth. Then we have
a dual representation for $\mathcal{M}_{\varepsilon}( o)$
according to the Fenchel-Legendre transformation.
\begin{proposition}\label{pro2}
The Fenchel-Legendre transformation of $\mathcal{M}_{\varepsilon}$ is:
\begin{equation*}\label{df_p}
\begin{array}{rl}
\mathcal{M}^{*}_{\varepsilon}(u)&:=\max\limits_{o}\left\{ ou-\mathcal{M}_{\varepsilon}(o)\right\}\\
&=\left\{
\begin{array}{lll}
\varepsilon (u\ln u+(1-u)\ln(1-u)),& u\in[0,1],\\
+\infty,& u\not\in[0,1].
\end{array}
\right.
\end{array}
\end{equation*}
In the  formulation above, we define $0\ln 0=0$.
\end{proposition}
\begin{proposition}\label{pro3}
The twice Fenchel-Legendre transformation of $\mathcal{M}_{\varepsilon}$ is:
\begin{equation*}
\mathcal{M}_{\varepsilon}^{**}(o)=\max\limits_{u\in[0,1]}\left\{ou-\varepsilon(u\ln u+(1-u)\ln(1-u))\right\}.
\end{equation*}
\end{proposition}

Since $\mathcal{M}_{\varepsilon}$ is convex and we have $\mathcal{M}_{\varepsilon}(o)=\mathcal{M}_{\varepsilon}^{**}(o)$.

On the other hand, the ReLU$(o)$ can be regarded as the maximum energy of K-means types clustering problem $\hat{u}=\underset{u\in[0,1]}{\arg\max}\{ou\}$.
It is easy to check $\hat{u}=1$ when $o\geqslant 0$ and $\hat{u}=0$ if $o< 0$. This is a binary segmentation to distinguish $o\geqslant 0$ and $o< 0$.

By smoothness, we have a soft thresholding segmentation according to the formulation of $\mathcal{M}_{\varepsilon}^{**}$ since it is also a smooth version of $\max$ function.
By changing the maximization problem to an equivalent minimization problem, we have that the sigmoid function $\mathcal{S}(o)$ is a minimizer of the following problem:
\begin{equation}\label{sigmoid_dual}
\underset{u\in[0,1]}{\min}\left\{-ou+\varepsilon(u\ln u+(1-u)\ln(1-u))\right\}
\end{equation}
when $\varepsilon=1$.

Compared to the variational segmentation method, the segmentation function $u$ in the above problem lacks of spatial regularization such as $u$ belongs to a bounded variation function space which can make the segmentation be piecewise constant and smooth. With this dual representation of the sigmoid
activation function, the convex shape prior and spatial regularization can be easily added to DCNNs through these variational models.

The second term in \eqref{sigmoid_dual} is an entropy term which forces $u$ to be smooth. The larger $\varepsilon$, the smoother $u$ is. It would be reduced to the binary segmentation (K-means) when $\varepsilon=0$. For DCNN layers, this entropy term is very helpful since it can make the back propagation to be stable.

\subsection{Convex Shape and Soft Thresholding Dynamic (CS-STD) With the Sigmoid Activation Function}
To add spatial regularization into DCNNs, we use an easily implementable regularization term called Thresholding Dynamic (TD) rather than TV. This regularization term is
\begin{equation}\label{eq:TD}
\mathcal{R}(\bm u)=\sqrt{\frac{\pi}{\sigma}}\sum_{\gamma=1}^{L-1}\int_{\Omega}u_{\gamma}(x) (k_{\sigma}*(1-u_{\gamma}))(x) \dx,
\end{equation}
where $k_{\sigma}$ is a Gaussian kernel with standard deviation $\sigma$. It has been shown \cite{Miranda2007} that $\mathcal{R}(\bm u) $
$\Gamma$-converges to $|\partial\Omega_{\gamma}|$ when $\sigma\rightarrow 0$.

The following Convex Shape and Soft Thresholding Dynamic (CS-STD) sigmoid segmentation can be easily derived according to the previous discussions:
\begin{equation}\label{eq:energy1}
 \widetilde{ u}=\underset{u\in[0,1]\bigcap\mathbb{C}}{\arg\min}\left\{
 \begin{array}{l}
 \underbrace{\langle- o, u\rangle+\varepsilon(u\ln u+(1-u)\ln(1-u))}_{:=\mathcal{F}(u;o)}\\
 +\lambda\underbrace{\langle e u, k_{\sigma}*(1- u)\rangle}_{:=\mathcal{R}( u)}
 \end{array}
 \right\}.
 \end{equation}
Here $\mathbb{C}$ is a convex shape condition set which is defined in (\ref{eq:C}), and it can guarantee $u$ to be a segmentation function of convex objects. The weighting function $e\geqslant 0$ is a given image edge detection function $e(x)=\frac{1}{1+||\nabla  v(x)||}$. It has been shown \cite{Wang2009} that $\mathcal{R}( u)\propto\sum_{\gamma=1}^{L-1}\int_{\partial \Omega_{\gamma}} e\mathrm{d}s$ when the kernel $k_{\sigma}$ satisfies some mild conditions. Thus $\mathcal{R}( u)$ is an active contour term which regularizes the length of the contours. 

\subsection{Our New Algorithm}
For the energy functional of (\ref{eq:energy1}), $\mathcal{F}$ is convex and $\mathcal{R}$ is concave when the kernel $k_{\sigma}$ is semi-positive definite. Thus the Difference of Convex Algorithm (DCA) \cite{Tao1997} can be applied. We can obtain an iteration algorithm: 
\begin{equation}\label{CSSTDiteration1}
\begin{array}{rl}
{{u}}^{t_1+1}&=\underset{u\in[0,1]\bigcap\mathbb{C}}{\arg\min}\left\{\mathcal{F}(u;o)+\lambda\mathcal{R}({u}^{t_1})+\lambda\langle  p^{t_1},{u}-{u}^{t_1} \rangle\right\}.\\
\end{array}
\end{equation}
Here $ p^{t_1}=(k_{\sigma}*(1- u^{t_1}))e-k_{\sigma}*(e u^{t_1})\in\partial \mathcal{R}({{u}}^{t_1})$ and $\partial \mathcal{R}({{u}}^{t_1})$ is the subgradient of the concave functional $\mathcal{R}$ at $ u^{t_1}$.
It can be shown that this iteration is energy descent and unconditionally stable without the condition set $\mathbb{C}$.

However, due to the existence of $\mathbb{C}$, the above
problem does not have a closed-form solution.
In order to get a sigmoid segmentation solver, we
use a pseudo projection algorithm to split this problem:
\begin{equation}\label{CSSTDiteration2}
\left\{
\begin{array}{rl}
u^{t_1+\frac{1}{2}}&=\underset{u}{\arg\min}\left\{\mathcal{F}(u;o)+\lambda\langle  p^{t_1},{u}\rangle\right\},\\
u^{t_1+1}&=\text{Proj}_{[0,1]\bigcap\mathbb{C}}(u^{t_1+\frac{1}{2}}).
\end{array}
\right.
\end{equation}
The first subproblem has an explicit solution formula and it solution is the regularized sigmoid solution:
\begin{equation}\label{regsigmoid}
u^{t_1+\frac{1}{2}}=\frac{1}{1+e^{\frac{-o+\lambda p^{t_1}}{\varepsilon}}}=\mathcal{S}(\frac{o-\lambda p^{t_1}}{\varepsilon}).
\end{equation}
Compared to the classic sigmoid, there are two improvements. Firstly, the dual variable $p^{t_1}$ can ensure that this sigmoid function has regularization effects to force the classification function to be nearly piece-wise constant. Secondly,
the entropy parameter $\varepsilon$ can guarantee that  the segmentation function is nearly binary, which enables us to  integrate the binary
convex shape condition in our method.

The second subproblem in \eqref{CSSTDiteration2} is to solve the following minimization problem
\begin{equation}\label{convexsubproblem}
u^{t_1+1}=\underset{u\in [0,1]\bigcap\mathbb{C}}{\arg\min}||u-u^{t_1+\frac{1}{2}}||^2.
\end{equation}
We use the active set method to solve this subproblem.
Let us analyze this problem. When $u$ fails to satisfy the condition $[0,1]\bigcap\mathbb{C}$, the condition should be activated. Then we have $(1-u(x))(g_r*(1-2u))(x)= 0$. It is easy to check $u=1$ always satisfies this condition. This observation leads us to the following simple active set iteration Algorithm \ref{alg1} in terms of proposition \ref{pro1}. Note we only require the quadratic convex shape condition to be satisfied for some given values for $r$. Surely, one can use more values of $r$ for other applications. We observe that 5 values of $r$ is enough for our testing cases.

Combining with the first subproblem, we summarize the CS-STD algorithm in Algorithm \ref{alg2}.

\begin{algorithm}
\caption{$\text{Proj}_{[0,1]\bigcap\mathbb{C}}$  for convex shapes}\label{alg1}
\KwIn{$u^{t_1+\frac{1}{2}}$. Different sphere radius $\bm r=(r_0,r_1,r_2,r_3,r_4)$.}
\textbf{Initialization:} $u^{0}=u^{t_1+\frac{1}{2}}$\\
\For{$t_2=0,1,\cdots$}{
1. Set $r=r_{\text{mod}(t_2,5)}$.\\
2. Find the active set
$$\mathbb{A}=\{x: (1-u^{t_2}(x))(g_r*(1-2u^{t_2}))(x)<0\}$$

2. Update
\begin{equation}
u^{t_2+1}(x)=\left\{
\begin{array}{rl}
1,&x\in\mathbb{A},\\
u^{t_2}(x),& x\notin\mathbb{A}.
\end{array}
\right.
\end{equation}
\\
3. Convergence check. If it is converged, end the algorithm.\\
}
%\Return Segmentation function $u^{t_1+1}=u^{t_2+1}$.
\KwOut{Segmentation function $u^{t_1+1}=u^{t_2+1}$ with convex shape.}
\end{algorithm}

\begin{algorithm}
\caption{CS-STD sigmoid activation function}\label{alg2}
\KwIn{The feature $ o$.}
\textbf{Initialization:} $u^0=\mathcal{S}( o).$\\
\For{$t_1=0,1,2,\cdots$}{
1. Compute the  solution of the first subproblem in \eqref{CSSTDiteration2} by regularized STD sigmoid \eqref{regsigmoid}.\\
%$$
%  u^{t_1+\frac{1}{2}}=\mathcal{S}\left(\frac{ o- \lambda p^{t_1}}{\varepsilon}\right).
%$$\\
2. Calculate the  pseudo projection $u^{t_1+1}=\text{Proj}_{[0,1]\bigcap\mathbb{C}}(u^{t_1+\frac{1}{2}})$ by Algorithm \ref{alg1}.\\
3. Convergence check. If it is converged, end the algorithm.\\
}
%\Return Segmentation function $u$.
\KwOut{Segmentation function $ u$ with convex shape prior.}
\end{algorithm}

\subsection{New CS-STD Sigmoid Block for DCNN }
We can use the general CS-STD sigmoid activation function as a block for some commonly used DCNNs and thus we can guarantee that the outputs of the new DCNNs to be smooth convex objects.  This can be done by unrolling
the Algorithm \ref{alg2} as some network layers. The original classification function sigmoid can be replaced by a variational problem which can handle convex convex prior, i.e. we replace the last layer of DCNNs by a variational problem and get
\begin{equation}\label{cs-std-cnn}
%\left\{
%\begin{array}{rl}
%\bm o^t=&\mathcal{A}^{t}\circ\mathcal{T}_{\bm \Theta^{t-1}}(\bm v^{t-1},\bm v^{t-2},\cdots,\bm v^0),t=1,\cdots, T.\\
%\bm v^{T}=&\underset{\bm u\in[0,1]\bigcap\mathbb{C}}{\arg\min}\{\mathcal{F}(\bm u;\bm o^{T})+\lambda\mathcal{R}(\bm u)\}.
%\end{array}
%\right.
\left\{
\begin{array}{l}
\left\{
\begin{array}{rl}
\bm o^t=&\mathcal{T}_{\bm \Theta^{t-1}}(\bm v^{t-1},\bm v^{t-2},\cdots,\bm v^0), t=1,\cdots, T,\\
\bm v^{t}=&\mathcal{A}^{t}(\bm o^t),t=1,\cdots, T-1,\\
\end{array}
\right.
\\
\\
\bm v^{T}=\underset{\bm u\in[0,1]\bigcap\mathbb{C}}{\arg\min}\{\mathcal{F}(\bm u;\bm o^{T})+\lambda\mathcal{R}(\bm u)\}.
\end{array}
\right.
\end{equation}
Theoretically, each activation function $\mathcal{A}^{t}$ appearing in \eqref{cs-std-cnn} can be replaced by a  regularized variational activation function and this could lead to vast variety of choices. To save computational sources, here we just replace the last layer to a  regularized one with convex shape prior.

\begin{figure*}
\includegraphics[width=1.01\linewidth]{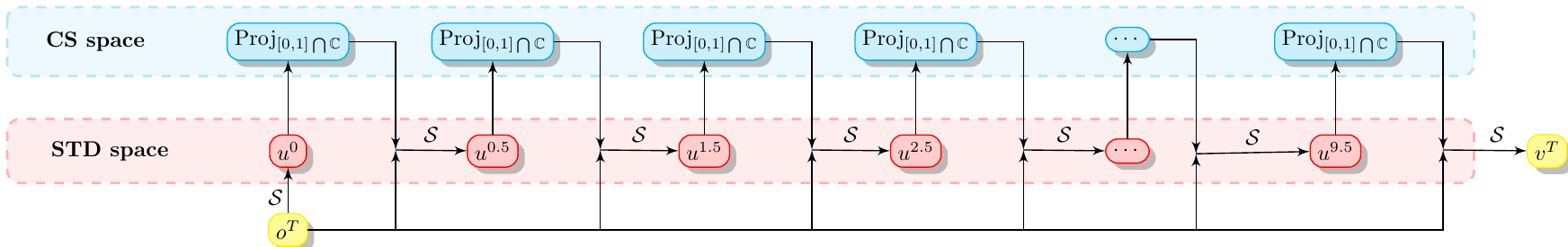}\caption{The CS-STD block unrolled by Algorithm \ref{alg2}. $\mathcal{S}$ stands for classical sigmoid operator. }\label{fig:CS-STD-net}
\end{figure*}

The second problem in \eqref{cs-std-cnn} needs to iteratively solved by Algorithm \ref{alg2}. Each iteration can be regarded as a DCNN layer, and thus Algorithm \ref{alg2} forms a new CS-STD sigmoid block for a new DCNN. To intuitively see the
information propagation among different spaces, we show a
network architecture schematic diagram in figure~\ref{fig:CS-STD-net}. In this figure, the red rectangle represents STD space in which the functions have piece-wise constants property, while the cyan rectangle is the Convex Shape (CS) space, $o^T$ is the features extracted by backbone network, and $v^T$ is the output of CS-STD block.

\subsection{Some Applications of the Proposed Methods}
In this section, we will show how to apply the proposed CS-STD block to the retinal images segmentation with some popular basic DCNNs such as DeepLabV3+.

\subsubsection{Sublevel set Representation}
\begin{figure}
\centering
\includegraphics[width=1.0\linewidth]{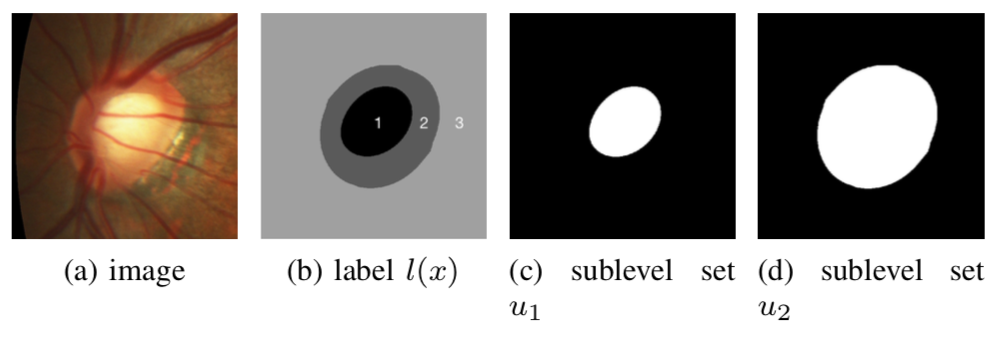}
%\subfloat[image]{\includegraphics[width=0.23\linewidth]{images/g0013.png}}~
%\subfloat[label $l(x)$]{\includegraphics[width=0.23\linewidth]{images/g0013_label.png}}~
%\subfloat[sublevel set $u_1$]{\includegraphics[width=0.23\linewidth]{images/g0013_u1.png}}~
%\subfloat[sublevel set $u_2$]{\includegraphics[width=0.23\linewidth]{images/g0013_u2.png}}~
\caption{The cup and disc areas in retinal images with sublevel set functions $u_1$ and $u_2$ representation can be both convex.}\label{fig:subleveset}
\end{figure}

In figure~\ref{fig:subleveset}, a local retinal image and the related ground truth are displayed in (a) and (b), respectively. The images need to be segmented into 3 phases for cup, disc and background. 
The ground truth label function $l(x)$ is shown in (b).
There are three important spatial priors for  retinal images segmentation. Firstly, the disc must contain the cup areas. Secondly, both of disc and cup should be convex. Thirdly, the
segmentation boundaries are smooth. In the next, we will show how to ensure the output of DCNNs can keep these spatial properties.

To keep the nested connection between cup and disc, the previously introduced sublevel set is adopted. In this application, the label function $l(x)\in\{1,2,3\}$, and the related sublevel sets $u_1$ and $u_2$ are displayed in figure~\ref{fig:subleveset} (c) and (d), respectively.
With this formulation, both objects represented by $u_1$ and $u_2$ are convex. Let us mention that the region labeled with $2$ would not be convex if we follow the indicative functions based segmentation method. The condition $\bm u=(u_1,u_2)\in\mathbb{U}$, where $\mathbb{U}$ is a sublevel functions set defined in \eqref{sublevelsetu}, can ensure the nested relationship of cup and disc. Therefore, the model \eqref{label_convex_discrete} is adopted in our CS-STD block. In \eqref{label_convex_discrete}, there are 3 classes features, denoted as $\hat{o}_1,\hat{o}_2,\hat{o}_3$. In fact, this classification criterion is the differences of these 3
classes features, i.e. $o_1=\hat{o}_1-\hat{o}_2$ and $o_2=\hat{o}_2-\hat{o}_3$. Therefore, in our method, the backbone network should find the difference features $o_1$ and $o_2$ for the 3 classes objects.

The smooth segmentation boundaries and the convex prior can be ensured by the CS-STD block. Thus we get a general
DCNN with the sublevel set representation as
\begin{equation}\label{cs-std-cnn-sublevel}
%\left\{
%\begin{array}{rl}
%\bm o^t=&\mathcal{A}^{t}\circ\mathcal{T}_{\bm \Theta^{t-1}}(\bm v^{t-1},\bm v^{t-2},\cdots,\bm v^0),t=1,\cdots, T.\\
%\bm v^{T}=&\underset{\bm u\in\mathbb{U}\bigcap\mathbb{C}}{\arg\min}\{\mathcal{F}(\bm u;\bm o^{T})+\lambda\mathcal{R}(\bm u)\}.
%\end{array}
%\right.
\left\{
\begin{array}{l}
\left\{
\begin{array}{rl}
\bm o^t=&\mathcal{T}_{\bm \Theta^{t-1}}(\bm v^{t-1},\bm v^{t-2},\cdots,\bm v^0), t=1,\cdots, T,\\
\bm v^{t}=&\mathcal{A}^{t}(\bm o^t), t=1,\cdots, T-1,\\
\end{array}
\right.
\\
\\
\bm v^{T}=\underset{\bm u\in\mathbb{U}\bigcap\mathbb{C}}{\arg\min}\{\mathcal{F}(\bm u;\bm o^{T})+\lambda\mathcal{R}(\bm u)\}.
\end{array}
\right.
\end{equation}
Here $\bm o^{T}=(o_1^T,o_2^T)$ is the difference of features for cup, disc and background. $\bm u=(u_1,u_2)$ is a vector-valued sublevel function. The only difference between \eqref{cs-std-cnn} and
\eqref{cs-std-cnn-sublevel} is the constraint $[0,1]$ which is replaced by $\mathbb{U}$. Since $\mathbb{U}$ is convex, the projection on it can be efficiently solved \cite{Liu2014}. Thus the algorithm for the second problem of \eqref{cs-std-cnn-sublevel} is almost the same with \eqref{cs-std-cnn}'s except for a projection step on $\mathbb{U}$. We do not plan to list the repetitive algorithm here.

\subsubsection{Backbone of DCNN}
\begin{figure*}
\includegraphics[width=1.02\linewidth]{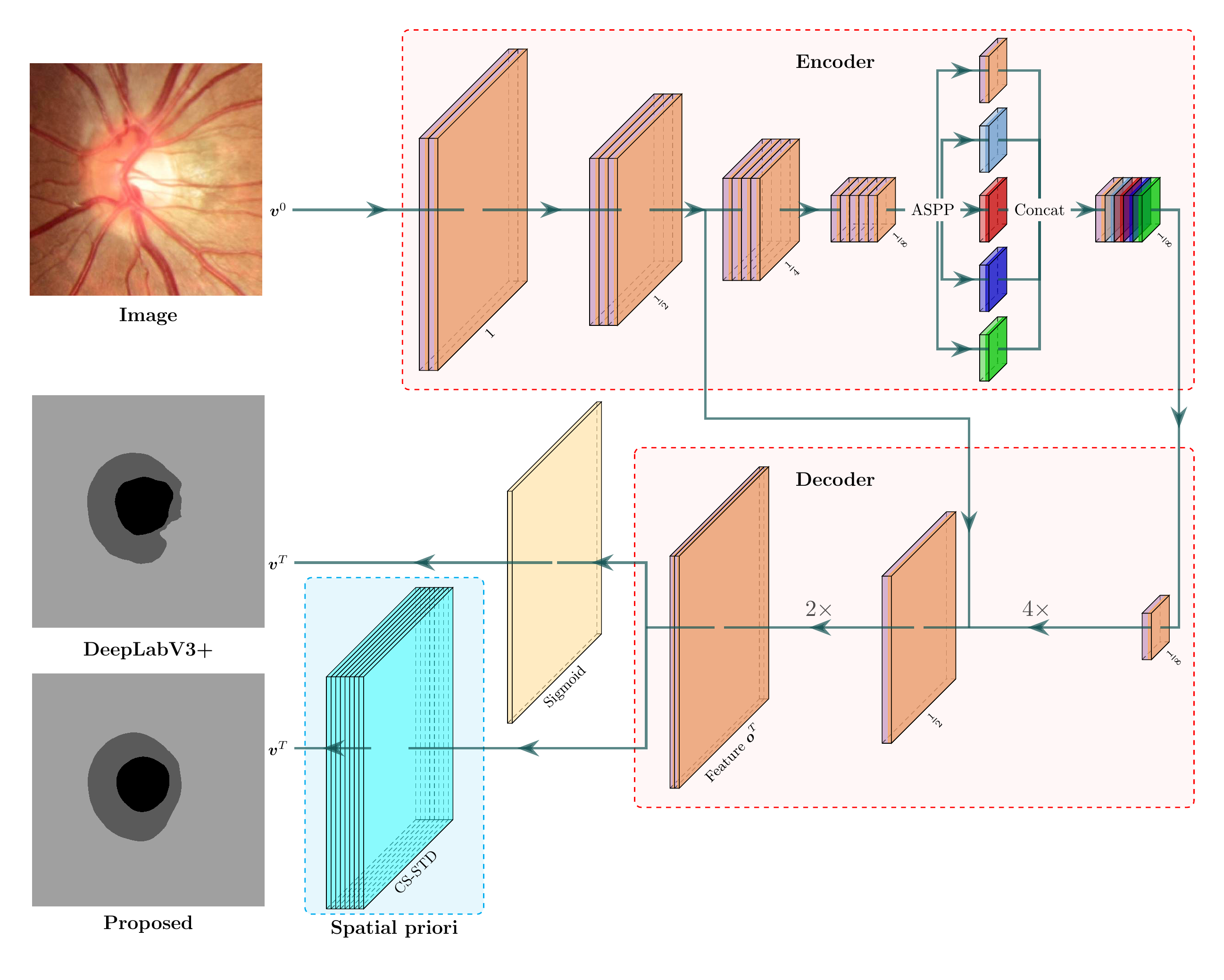}\caption{The architecture of the CS-STD based DeeplabV3+ segmentation network.}\label{fig:netarch}
\end{figure*}

To test our algorithm on DCNNs, we adopt the DeeplabV3+
encoder-decoder structure to extract features $\bm o^T$.
For a small number of parameters and fast implementation, the  MobileNetV2 backbone \cite{Sandler2018} is applied.
The details of the whole network are displayed in figure~\ref{fig:netarch}.
The input image is denoted as $\bm v^0$. After several convolutions, inverted residual blocks and $4$ times downsampling with rate $0.5$, the ASPP is adopted. In the decoder part, two interpolation  operators with upsampling rates $4$ and $2$  are employed to restore the resolution of the features. As mentioned earlier, the feature $\bm o^T$ in this network would be the differences between features for cup, disc and background. Then, the CS-STD block is placed after $\bm o^T$ feature layer, and it would produce convex disc and cup objects represented by two sublevel set functions $u_1$ and $u_2$. The output label function $l(x)$ can be obtained according to \eqref{eq:sublevel2label} when we get these two sublevel set functions.

It is observed that the proposed CS-STD is a plug-and-play
block. Let regularization parameter $\lambda=0$, entropy parameter $\varepsilon=1.0$. By removing convex prior pseudo projection, it would reduce to the classic sigmoid activation function. On the other hand, if one removes this block in the training step and restores it in the prediction, it would be equivalent to a post-processing method. The proposed CS-STD block can work on any semantic segmentation DCNNs if the datasets have convex prior.

\subsubsection{Loss Function}
Since the convex prior set $\mathbb{C}$ is composed of binary functions, thus in the CS-STD sigmoid, the entropy parameter $\varepsilon$ should be very small, e.g. $\varepsilon=0.1$. Therefore,
the output of the DCNN would be nearly binary, and thus the cross entropy loss function cannot be directly applied because $\ln 0$ would be $\infty$. We adopted the smooth Dice loss \cite{milletari2016v}
$$\mathcal{L}(\bm v^T, \bm u_{\text{true}})=1-\frac{2\langle \bm v^T, \bm u_{\text{true}}\rangle}{||\bm v^T||^2+||\bm u_{\text{true}}||^2}$$
to train the DCNN, where $\bm u_{\text{true}}$ is the sublevel set of ground truth.

\section{Numerical Experimental Results}\label{experiment}
In this section, we shall first design a very simple experiment
to show the intuition of the proposed CS-STD algorithms. Next, we will evaluate the performance of CS-STD block on a dataset. Then the generalization ability of
shape prior in DCNNs is shown. Finally, the robustness with noise for the proposed method  will be tested.

\subsection{The Performance of CS-STD}
\begin{figure}%[htbp]
\centering
\includegraphics[width=1.0\linewidth]{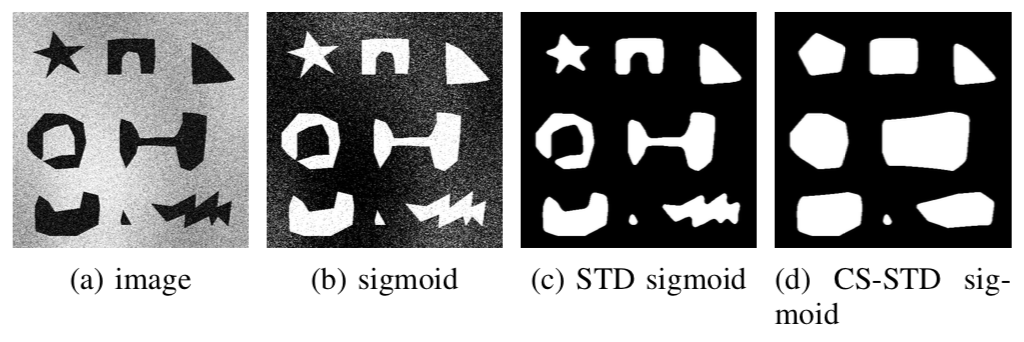}
%\subfloat[image]{\includegraphics[width=0.24\linewidth]{images/image.png}}~
%\subfloat[sigmoid]{\includegraphics[width=0.24\linewidth]{images/u_sigmoid.png}}~
%\subfloat[STD sigmoid]{\includegraphics[width=0.24\linewidth]{images/u_STD.png}}~
%\subfloat[CS-STD sigmoid]{\includegraphics[width=0.24\linewidth]{images/u_CSSTD.png}}~
\caption{The comparison of sigmoid with and without spatial regularization and convex shape prior.}\label{fig:alg2}
\end{figure}

In this subsection, we test the performance of Algorithm \ref{alg2}. In figure~\ref{fig:alg2}, there is an image $v(x)$ which contains several simple geometry objects. We will show our method can
get multiple convex objects by using one classification function. 
According to the sublevel set representation, we need to find the difference between the features of the objects and background.
We simply choose the region variance as the feature $\hat{o}_{\gamma}=-||v-\mu_{\gamma}||^2 $ for $\gamma=1,2$. Here $\mu_{\gamma}$ are given means of the gray values of objects and background, respectively.
As mentioned earlier, the difference of features $o=\hat{o}_1-\hat{o}_2$. We put $o$ as the input of Algorithm \ref{alg2} and show the segmentation results in figure~\ref{fig:alg2} (d). For comparison, we give the results produced by sigmoid segmentation and STD sigmoid only with regularization term $\mathcal{R}$ in figure~\ref{fig:alg2} (b) and (c). One can see that the STD sigmoid can make the segmentation piece-wise constants and  CS-STD sigmoid provides convex objects with smooth boundaries. In this experiment, the parameters in Algorithm \ref{alg2} are set as $\lambda=10, \bm r=(15,10,5,3,1), \varepsilon=0.05$.
The outer iteration of Algorithm \ref{alg2} will converge within $t_1=10$ for most of the cases, while the inner iteration for Algorithm \ref{alg1} depends on the non-convexity of the objects, usually, $100$ iterations is sufficient. In the experiment, we choose outer iteration number $t_1=10$ and inner iteration number $t_2=50$.

\begin{figure}%[htbp]
\centering
\includegraphics[width=1.0\linewidth]{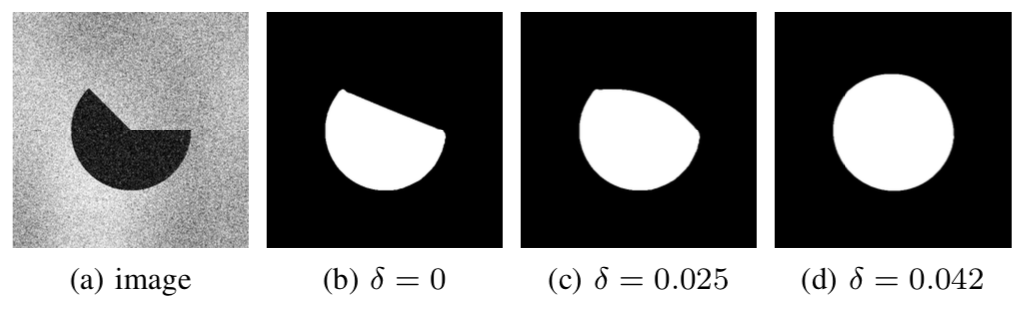}
%\subfloat[image]{\includegraphics[width=0.24\linewidth]{images/image_1.png}}~
%\subfloat[$\delta=0$]{\includegraphics[width=0.24\linewidth]{images/u_STD_0.png}}~
%\subfloat[$\delta=0.025$]{\includegraphics[width=0.24\linewidth]{images/u_STD_1.png}}~
%\subfloat[$\delta=0.042$]{\includegraphics[width=0.24\linewidth]{images/u_STD_2.png}}~
\caption{CS-STD sigmoid Algorithm \ref{alg2} with different curvature constraints.}\label{fig:alg1}
\end{figure}

We observe that the active set $\mathbb{A}$ in Algorithm \ref{alg1} is an approximated curvature to measure the degree of curves bending. An interesting thing is that the Algorithm \ref{alg1} can be extended to force the curvatures $\kappa$ of the convex object boundaries larger than a given value. This can be easily done by setting
$$\mathbb{A}=\{x: (1-u^{t_2}(x))(g_r*(1-2u^{t_2}))(x)<\delta\}$$
in the Algorithm \ref{alg1}. Here $\delta\geqslant 0$ is a given value which is related to the curvature $\kappa$ of object boundaries. We design a toy experiment to show this.
In figure~\ref{fig:alg1}, we show the segmentation results of the CS-STD Algorithm \ref{alg2} with different
$\delta$ values in active set $\mathbb{A}$. The feature $o$ of the  image is obtained as the same as in figure~\ref{fig:alg2}. The parameters are set as the same as the previous experiment except for $\bm r=(25,25,25,25,1), t_1=20$ for fast convergence.
As can be seen from this figure, though the segmentations are all convex, they are very different. The straight lines ($\kappa=0$) can be allowed for the boundaries of the convex objects when $\delta=0$.
When $\delta$ increases, the segmented convex object
goes to a circle gradually. This means that our algorithm not only can ensure the segmentations are convex, but also can provide special convex shapes such as circles according to different approximate curvature constraints.

\subsection{Evaluation on REFUGE Challenge Data Set}
\begin{figure*}
\centering
\includegraphics[width=1.0\linewidth]{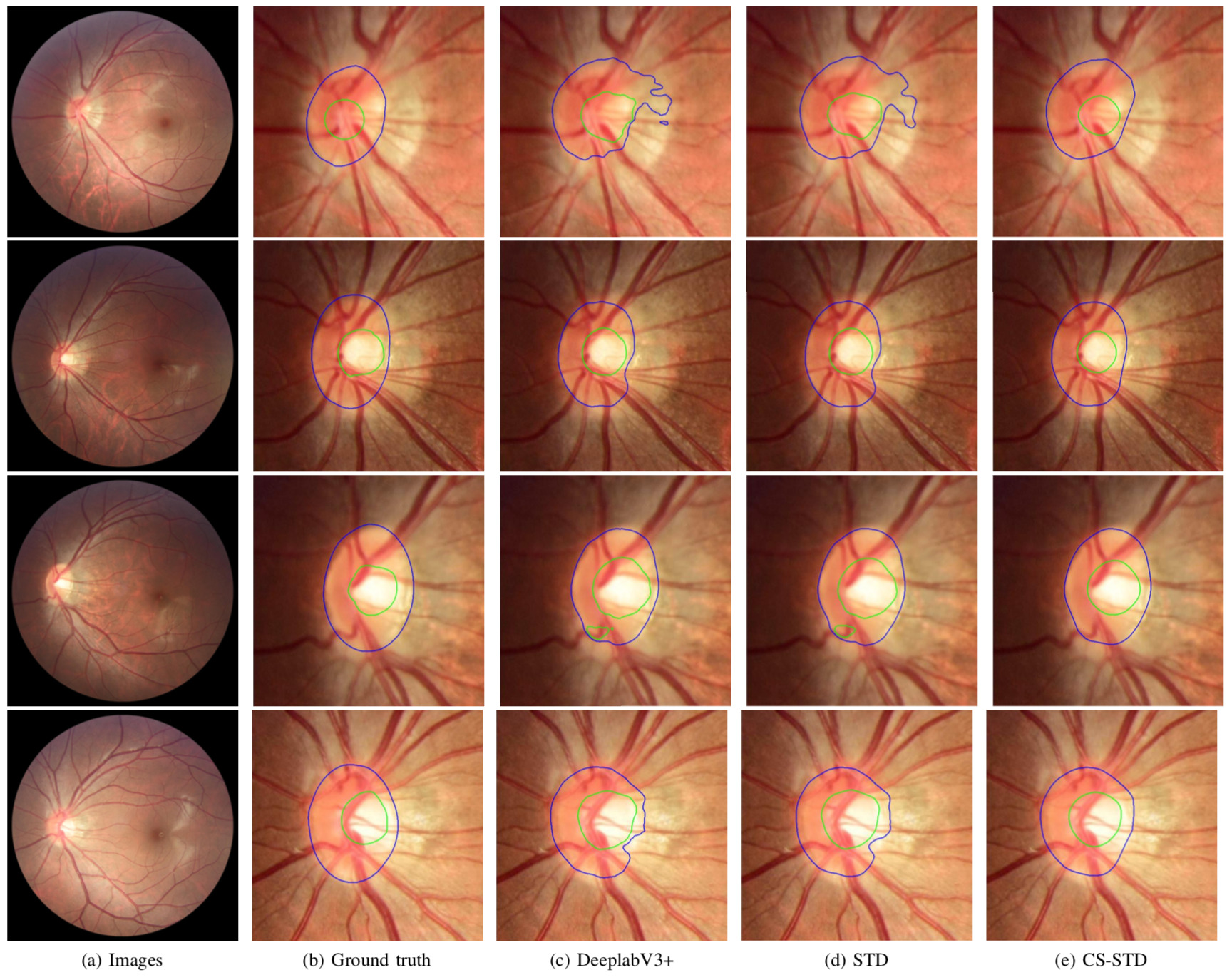}
%{\includegraphics[width=0.19\linewidth]{figures/ori_T0135.jpg}}~
%{\includegraphics[width=0.19\linewidth]{figures/gt_T0135.jpg}}~
%{\includegraphics[width=0.19\linewidth]{figures/deeplab_T0135.jpg}}~
%{\includegraphics[width=0.19\linewidth]{figures/std_T0135.jpg}}~
%{\includegraphics[width=0.19\linewidth]{figures/csstd_T0135.jpg}}~
%\\
%\vspace{0.05cm}
%{\includegraphics[width=0.19\linewidth]{figures/ori_T0308.jpg}}~
%{\includegraphics[width=0.19\linewidth]{figures/gt_T0308.jpg}}~
%{\includegraphics[width=0.19\linewidth]{figures/deeplab_T0308.jpg}}~
%{\includegraphics[width=0.19\linewidth]{figures/std_T0308.jpg}}~
%{\includegraphics[width=0.19\linewidth]{figures/csstd_T0308.jpg}}~
%\\
%\vspace{0.05cm}
%{\includegraphics[width=0.19\linewidth]{figures/ori_T0340.jpg}}~
%{\includegraphics[width=0.19\linewidth]{figures/gt_T0340.jpg}}~
%{\includegraphics[width=0.19\linewidth]{figures/deeplab_T0340.jpg}}~
%{\includegraphics[width=0.19\linewidth]{figures/std_T0340.jpg}}~
%{\includegraphics[width=0.19\linewidth]{figures/csstd_T0340.jpg}}~
%\\
%\vspace{-0.3cm}
%\subfloat[Images]{\includegraphics[width=0.19\linewidth]{figures/ori_T0195.jpg}}~
%\subfloat[Ground truth]{\includegraphics[width=0.19\linewidth]{figures/gt_T0195.jpg}}~
%\subfloat[DeeplabV3+]{\includegraphics[width=0.19\linewidth]{figures/deeplab_T0195.jpg}}~
%\subfloat[STD]{\includegraphics[width=0.19\linewidth]{figures/std_T0195.jpg}}~
%\subfloat[CS-STD]{\includegraphics[width=0.19\linewidth]{figures/csstd_T0195.jpg}}~
\caption{Visual quality of the sigmoid, STD-sigmoid, CS-STD-sigmoid on Refuge test set. The basic network is DeeplabV3+ with backbone MobileNetV2.}\label{fig:refuge-test}
\end{figure*}

\begin{figure*}
\centering
\includegraphics[width=1.0\linewidth]{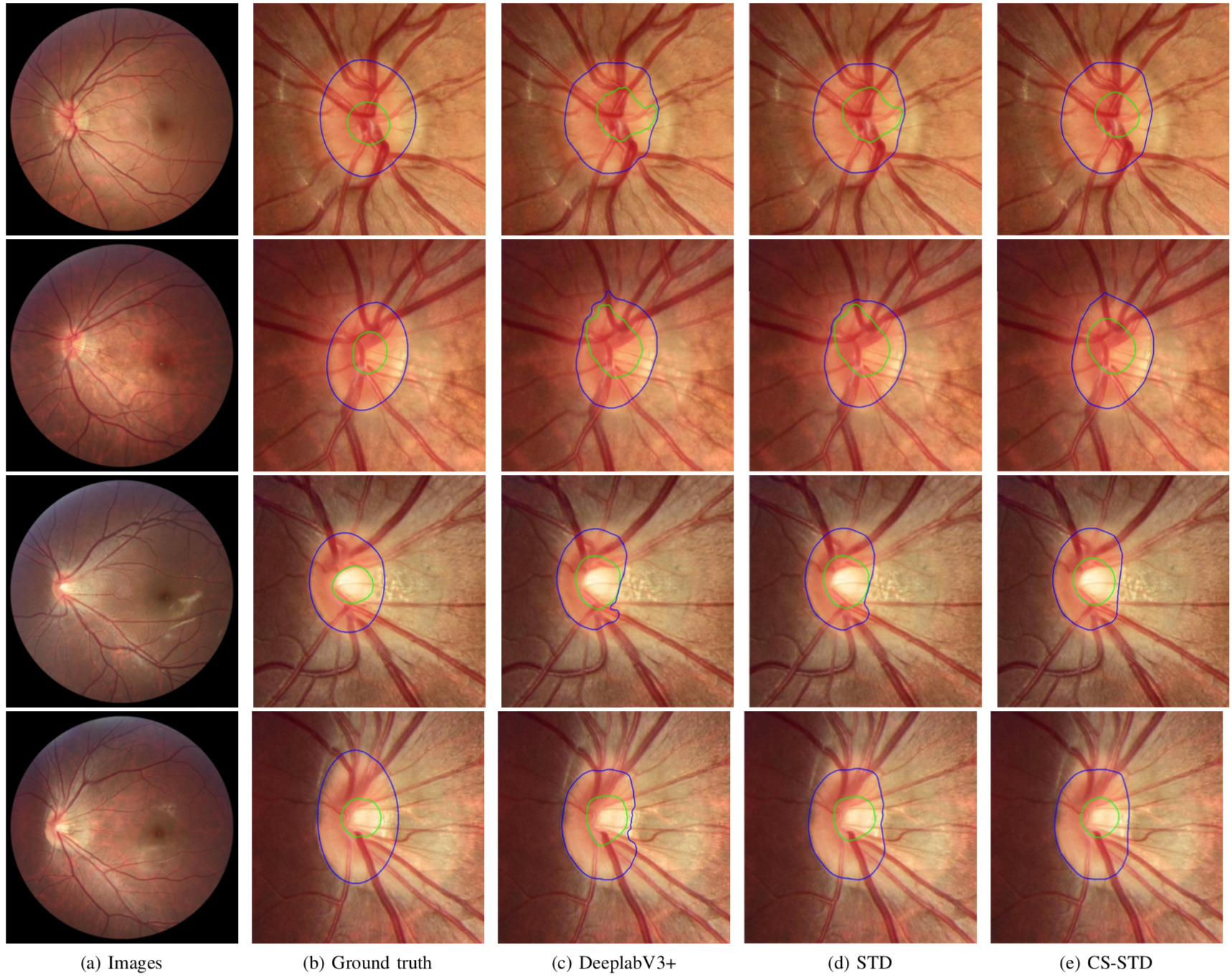}
%{\includegraphics[width=0.19\linewidth]{figures/ori_V0256.jpg}}~
%{\includegraphics[width=0.19\linewidth]{figures/gt_V0256.jpg}}~
%{\includegraphics[width=0.19\linewidth]{figures/deeplab_V0256.jpg}}~
%{\includegraphics[width=0.19\linewidth]{figures/std_V0256.jpg}}~
%{\includegraphics[width=0.19\linewidth]{figures/csstd_V0256.jpg}}~
%\\
%\vspace{0.05cm}
%{\includegraphics[width=0.19\linewidth]{figures/ori_V0273.jpg}}~
%{\includegraphics[width=0.19\linewidth]{figures/gt_V0273.jpg}}~
%{\includegraphics[width=0.19\linewidth]{figures/deeplab_V0273.jpg}}~
%{\includegraphics[width=0.19\linewidth]{figures/std_V0273.jpg}}~
%{\includegraphics[width=0.19\linewidth]{figures/csstd_V0273.jpg}}~
%\\
%\vspace{0.05cm}
%{\includegraphics[width=0.19\linewidth]{figures/ori_V0301.jpg}}~
%{\includegraphics[width=0.19\linewidth]{figures/gt_V0301.jpg}}~
%{\includegraphics[width=0.19\linewidth]{figures/deeplab_V0301.jpg}}~
%{\includegraphics[width=0.19\linewidth]{figures/std_V0301.jpg}}~
%{\includegraphics[width=0.19\linewidth]{figures/csstd_V0301.jpg}}~
%\\
%\vspace{-0.3cm}
%\subfloat[Images]{\includegraphics[width=0.19\linewidth]{figures/ori_V0304.jpg}}~
%\subfloat[Ground truth]{\includegraphics[width=0.19\linewidth]{figures/gt_V0304.jpg}}~
%\subfloat[DeeplabV3+]{\includegraphics[width=0.19\linewidth]{figures/deeplab_V0304.jpg}}~
%\subfloat[STD]{\includegraphics[width=0.19\linewidth]{figures/std_V0304.jpg}}~
%\subfloat[CS-STD]{\includegraphics[width=0.19\linewidth]{figures/csstd_V0304.jpg}}~
\caption{Visual quality of the sigmoid, STD-sigmoid, CS-STD-sigmoid on Refuge validation set. The basic network is DeeplabV3+ with backbone MobileNetV2.}\label{fig:refuge-val}
\end{figure*}

In this section, we test the performance of CS-STD block
on DCNNs.
The Refuge challenging data set contains $400$ train, $400$ validation, and $400$ test images. The image size is $ 2056\times 2124$. The disc, cup and background regions
should be segmented for further diagnosis of glaucoma. Since the vertical cup to disc ratio is a very important diagnostic index, the segmentation of disc and cup plays a very key role in this process.
To get a suitable image size, we extract a region of interested (ROI) with size $512\times512$ in each image
 for training and testing. This can be done with a rough pre-training U-net. We use the $400$ train images for train set, and then apply the trained network to predict the segmentation on $400$ test and $400$ validation images.
The batch size of train is $4$ and the total train epoch is $100$. As for the train rate, we set it as $2.5e-5$  and reduce it using the polynomial decay with a power of 0.9 as mentioned in \cite{v2}. In the STD based block, the parameters
are set as $\varepsilon=0.1, t_1=10, \bm r=(15,15,15,15,1), \bm \lambda=(5.0,10.0)$. To accelerate the train, we let $t_2=1$ in the train and reset it as $t_2=50$ in the prediction.
The dice measure (DM)
$$DM= \frac{2N_{TP}}{2N_{TP}+N_{FN}+N_{FP}}\times 100 \%$$
is adopted to evaluate the accuracy of the segmentation.
Here $N_{TP},N_{FN}$ and $N_{FP}$ are the number of
true positive, false negative, and false positive pixels, respectively. For comparison, we report the DM of DeepLabV3+ (without spatial smoothness and convexity), STD (with spatial smoothness but without convexity), CS-STD (with both spatial smoothness and convexity) in table~\ref{tab1}.
We also take a recent pOSAL-seg method for this task in \cite{Wang2019} for comparison, in which a morphology-aware loss
is proposed to force the segmentation to be smooth.
It can be observed that the proposed convex prior can improve the accuracy of the segmentation, especially for cup region. It can improve $3.4\% DM$ for cup region on the test set. On the other hand, let us notice the visual effects for these methods. In figure~\ref{fig:refuge-test} and \ref{fig:refuge-val}, we show the part results of the segmentation results. In these figures, the blue and green lines are the boundaries of disc and cup regions, respectively. As can be seen from these two figures, the results of DeeplabV3+ cannot ensure the disc and cup regions are both smooth and convex, but our CS-STD can provide convex disc and cup objects with smooth boundaries.

\begin{table}[htp]
\caption{ $DM$ values of different methods for Refuge validation and test sets. }\label{tab1}
\begin{center}
\begin{tabular}{cccccccc}
\toprule[2pt]
&&&val. &set&&test &set\\
\cline{4-5}\cline{7-8}
&methods&&disc&cup&&disc&cup\\
\midrule[2pt]
\multirow{2}*{Existing}
&STD\cite{Liu2020}&&95.1&86.7&&95.2&85.0\\
&pOSAL-seg\cite{Wang2019}&&93.2&86.9&&-&-\\
\midrule[2pt]
\multirow{1}*{Baseline}&DeeplabV3+\cite{v3+}&&95.0&86.4&&95.1&84.3\\
\midrule[2pt]
\multirow{1}*{Proposed}
&CS-STD&&\textbf{95.1}&\textbf{88.3}&&\textbf{95.2}&\textbf{87.7}\\

%\midrule[2pt]
%\multirow{3}*{Existing}&DeepLabV3+(0 iter.)\cite{v3+}&82.20\\
%\cline{2-3}
%&DeepLabV3+(50 K iter.)\cite{v3+}&91.34\\
%
%&VP-TV-DeepLabV3+(50 K iter.)\cite{Li2019}&93.41\\
%\midrule[2pt]
%\multirow{2}*{Proposed}&STD-DeepLabV3+(50 K iter.)&94.40\\
%\cline{2-3}
%&VP-STD-DeepLabV3+(50 K iter.)&\textbf{95.21}\\
\bottomrule[2pt]
\end{tabular}
\end{center}
\end{table}

\subsection{Generalization Ability on RIM-ONE-r3 Data set}
It is well-known that the generalization ability is a big problem for the DCNNs. That is, a trained DCNNs just can work well on the data which are similar (e.g. train and test data obey the same distribution), and it would rapidly degrade when applying it to a new dataset.
In this section, we will show that
the spatial prior in DCNNs can improve the generalization ability of DCNNs if the images have smoothness and convex shape prior. We first train the DeeplabV3+ and proposed CS-STD methods on Refuge train set, and then apply it to predict a new retinal image dataset called RIM-ONE-r3 \cite{Fumero2011}. This dataset contains $60$ test images whose $rgb$ values are totally different from the Refuge dataset used in the previous section. We list the $DM$ results for DeeplabV3+ and our CS-STD based methods in table~\ref{tab2}. To compare the domain adaptation method which is designed to address this problem, we also list some results of this kind method in table~\ref{tab2}.
It can be observed that the DM values of DeeplabV3+ are degraded rapidly due to the large difference between train and test data. The GAN based domain adaptation methods can partly prevent this degradation and improve the DM values. When we use the convex shape prior CS-STD, the segmentation of disc and cup can still be smooth and convex. Thus it can also provide good results. To get
these results, we use a large regularization parameter $\bm \lambda=(200,30)$ for disc and cup, and choose the smoothness control kernel in STD $k_{\sigma}=65$ to ensure that the segmentation results are smooth enough. It means that one can use different regularization parameters in train and prediction steps according to different application requirements. Let us mention that we do not use any post-processing technique while other methods may adopt morphological post-processing method to keep the segmentation to be smooth. Besides, our method does not use any domain adaptation technique,
so one can integrate our method with the domain adaptation method to further improve the generalization ability for retinal images segmentation.
\begin{table}[htp]
\caption{ $DM$ values of different methods for training on Refuge train set and predicting on RIM-ONE-r3 test set. }\label{tab2}
\begin{center}
\begin{tabular}{ccccc}
\toprule[2pt]
&methods&&disc&cup\\
\midrule[2pt]
\multirow{4}*{Existing}
&TD-GAN\cite{Zhang2018}&&85.3&72.8\\
&Hoffman \emph{et al.} \cite{Hoffman2016}&&85.2&75.5\\
&Javanmardi \emph{et al.} \cite{Javanmardi2018}&&85.3&77.9\\
&pOSAL\cite{Wang2019}&&86.5&78.9\\
\midrule[2pt]
\multirow{1}*{Baseline}&DeeplabV3+\cite{v3+}&&85.4&70.9\\
\midrule[2pt]
\multirow{1}*{Proposed}
&CS-STD&&\textbf{92.2}&\textbf{80.7}\\
\bottomrule[2pt]
\end{tabular}
\end{center}
\end{table}
\begin{figure*}
\hspace{0.5cm}
\centering
\includegraphics[width=1.0\linewidth]{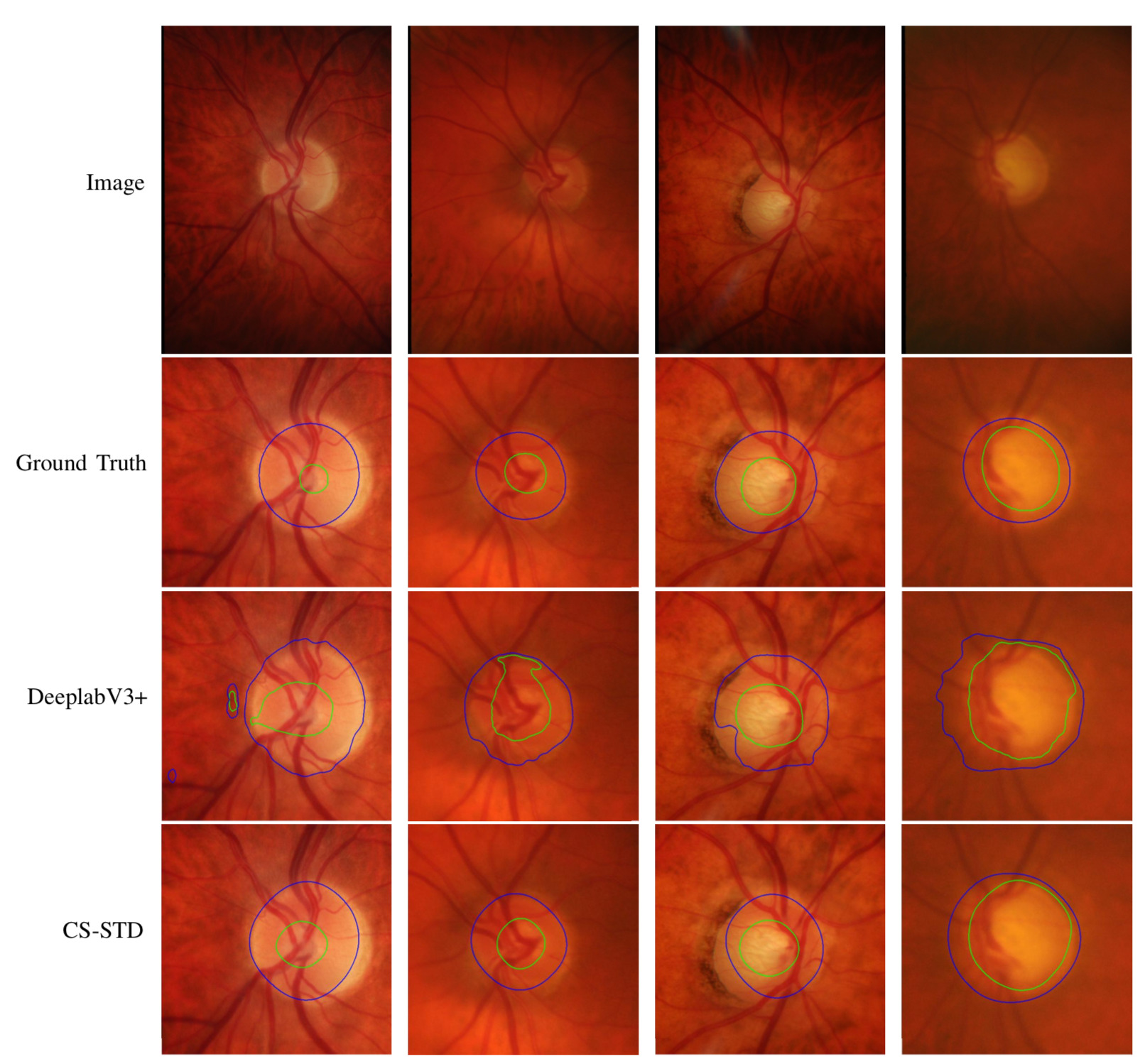}

%{\includegraphics[width=0.19\linewidth]{figures/ori_N-8-L.jpg}}~
%{\includegraphics[width=0.19\linewidth]{figures/ori_G-4-L.jpg}}~
%{\includegraphics[width=0.19\linewidth]{figures/ori_G-33-R.jpg}}~
%{\includegraphics[width=0.19\linewidth]{figures/ori_G-39-L.jpg}}~
%\put(-445,70){Image}
%\\
%\vspace{0.05cm}
%\hspace{0.5cm}
%{\includegraphics[width=0.19\linewidth]{figures/gt_N-8-L.jpg}}~
%{\includegraphics[width=0.19\linewidth]{figures/gt_G-4-L.jpg}}~
%{\includegraphics[width=0.19\linewidth]{figures/gt_G-33-R.jpg}}~
%{\includegraphics[width=0.19\linewidth]{figures/gt_G-39-L.jpg}}~
%\put(-475,50){Ground Truth}
%\\
%\vspace{0.05cm}
%\hspace{0.5cm}
%{\includegraphics[width=0.19\linewidth]{figures/deeplab_N-8-L.jpg}}~
%{\includegraphics[width=0.19\linewidth]{figures/deeplab_G-4-L.jpg}}~
%{\includegraphics[width=0.19\linewidth]{figures/deeplab_G-33-R.jpg}}~
%{\includegraphics[width=0.19\linewidth]{figures/deeplab_G-39-L.jpg}}~
%\put(-470,50){DeeplabV3+}
%\\
%\vspace{0.05cm}
%\hspace{0.5cm}
%{\includegraphics[width=0.19\linewidth]{figures/csstd_N-8-L.jpg}}~
%{\includegraphics[width=0.19\linewidth]{figures/csstd_G-4-L.jpg}}~
%{\includegraphics[width=0.19\linewidth]{figures/csstd_G-33-R.jpg}}~
%{\includegraphics[width=0.19\linewidth]{figures/csstd_G-39-L.jpg}}~
%\put(-455,50){CS-STD}
\caption{Visual quality of training on Refuge train set and predicting on RIM-ONE-r3 test set.}\label{fig:rimone-test}
\end{figure*}

\subsection{Robustness for Noise}
\begin{figure*}[tb!] 
\begin{minipage}{0.55\linewidth}
%\begin{table}[htp]
%\caption{ $DM$ values of DeepLabV3+ and CS-STD for cup regions in the Refuge test sets (400 images) under different levels Gaussian noise. }\label{tab3}
%\begin{center}
\begingroup
\setlength{\tabcolsep}{3.5pt} % Default value: 6pt
\renewcommand{\arraystretch}{1.5}
\small
\begin{tabular}{cccccccccc}
\toprule[2pt]
Noise levels&&&&&$\sigma$&&&&\\
\cline{2-10}
&0&1&2&3&4&5&6&7&8\\
\hline
%\midrule[2pt]
DeepLabV3+\cite{v3+}&84.3&
84.2&84.1&83.8&83.6&83.1& 82.6& 82.4& 81.7\\
CS-STD&87.7&87.6&87.5&87.3&86.9&86.6&86.2&85.9&85.5\\
\midrule[2pt]
Noise levels&&&&&$\sigma$&&&&\\
\cline{2-10}
&9&10&11&12&13&14&15&16&17\\
\hline
DeepLabV3+\cite{v3+}& 81.2&80.5&80.1&79.5&79.1&78.7&77.8&77.4&76.6\\
CS-SCT&85.1&84.6&84.0&83.5&83.2&82.7&82.3&81.3&81.2\\
\midrule[2pt]
Noise levels&&&&&$\sigma$&&&&\\
\cline{2-10}
&18&19&20&21&22&23&24&25&\\
\hline
DeepLabV3+\cite{v3+}&76.1&75.3&74.6&73.9&72.8&71.6&70.5&69.0\\
CS-STD&81.2&79.9&79.8&78.9&77.7&76.9&75.5&74.7\\
\bottomrule[2pt]
\end{tabular}
\endgroup
%\end{center}
%\end{table}
\end{minipage}
\begin{minipage}{0.45\linewidth}
\vspace{0.2cm}
\centering
\includegraphics[width=1.0\linewidth]{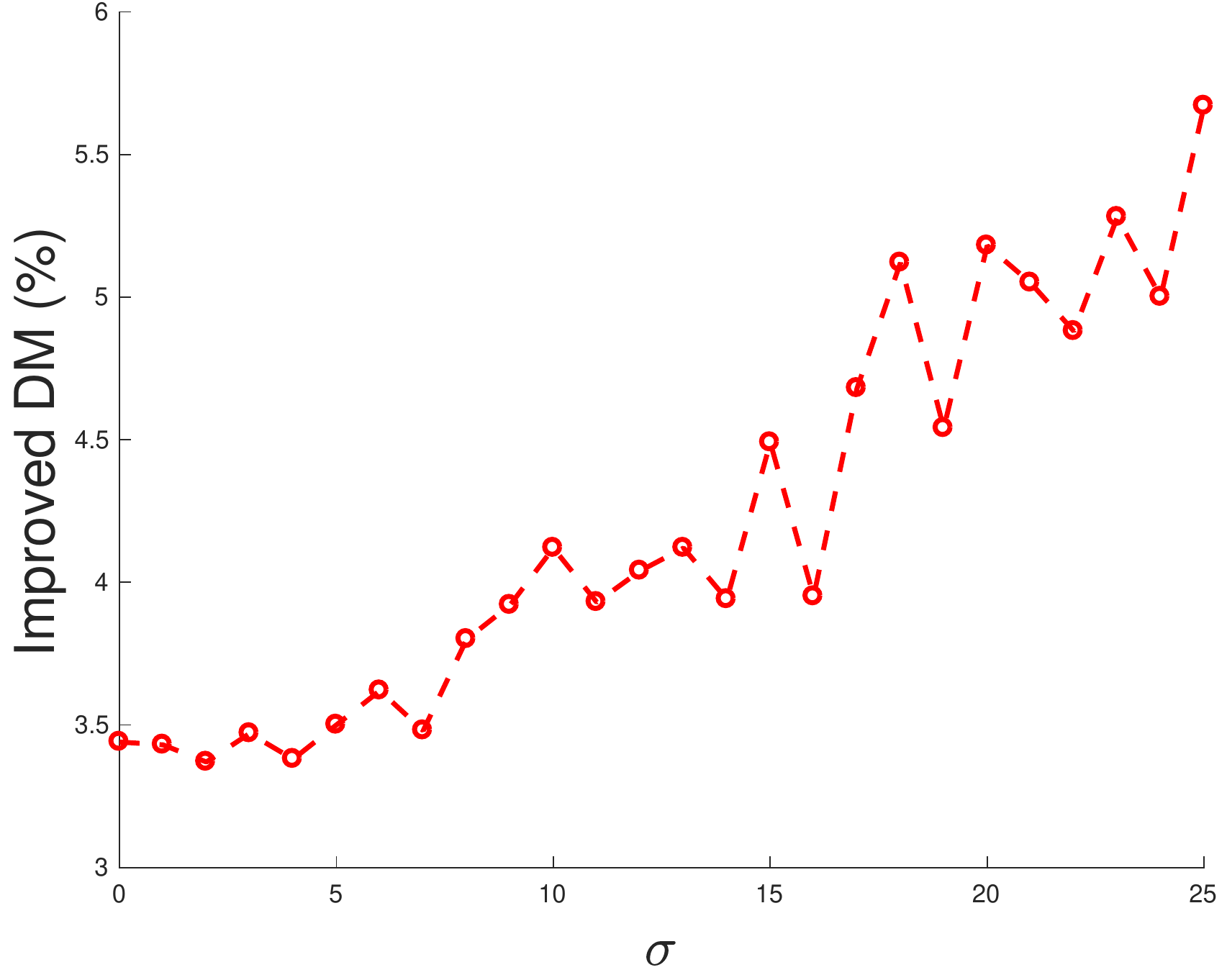}
\end{minipage}
\caption{The improved DM values for cup regions in the Refuge test sets (400 images) between DeeplabV3+ and the proposed CS-STD under different levels of noise with standard deviation $\sigma$. }\label{fig:noise_DM}
\end{figure*}

To show the spatial prior can improve the robustness
performance of the DCNN under noise, we first train the DeeplabV3+ and CS-STD based DCNN on the noise free train data of REFUGE, and then add some Gaussian noise to the $400$ test images of REFUGE with different standard deviations $\sigma$ from $0$ to $25$. The DM indexes of cup regions for DeeplabV3+ and the proposed CS-STD based method are listed in left table in figure \ref{fig:noise_DM}. It can be observed that the DM values can be improved by our spatial smoothness and convex prior in all the cases. It has $3.4\%$ improvement
under the noise free test data,  and reaches $5.7\%$ when the noise level is increased to $\sigma=25$. We plot the differences of DM values between DeeplabV3+ and CS-STD under the noise with standard deviations $\sigma\in\{0,1,2,
\cdots,25\}$
in the right of figure~\ref{fig:noise_DM}. It can be seen that the related curve has upward tendency, which indicates that the 
spatial prior can improve the robustness of the segmentation method.

\section{Conclusion and Discussion}
We have proposed a general framework for DCNN with convex shape prior. By explaining the activation function
as a soft classification function which is related to a variational problem, one can easily add the convex shape prior to existing DCNN architecture. We also show the intrinsic connections of sigmoid activation functions and the classic variational based image segmentation models. The proposed dual segmentation method can integrate many successful techniques in variational based image segmentation into DCNNs. We also show the application of our method on optic disc and cup of eye images to demonstrate the efficiency of the proposed method by numerical experiments.

In the current implementation, the regularization  and entropic parameters are both fixed. In fact, they both could be learned as well. Besides, the kernels in STD also can be learned, which may lead to some different regularization constraints rather than smooth boundaries.
Another possible extension is to use similar ideas on the activation function ReLU. ReLU can be also regularized by our method. This may be beneficial for extracting entire piece-wise constant  features. We will  work on these aspects in some future works.

\bibliography{reference}
\bibliographystyle{IEEEtran}

\end{document}